\documentclass[twocolumn]{article}
\usepackage[top=2cm, bottom=2.5cm, left=2cm, right=2cm]{geometry}
\usepackage{times,textcomp}
\usepackage[T1]{fontenc}
\usepackage[utf8]{inputenc}
\usepackage{amssymb,amsmath}
\usepackage{enumitem}
\usepackage{eurosym}
\usepackage{listings}
\usepackage{float,caption}
\usepackage{diagbox}
\usepackage{tabularx}

\usepackage[giveninits=true]{biblatex}
\setlist[itemize]{itemsep=0pt}

\usepackage{titlesec} 
\titleformat{\section}{\large\bfseries}{\thesection}{1em}{\MakeUppercase} 
\titleformat{\subsection}{\large\bfseries}{\thesubsection}{1em}{}
\titlespacing*{\section}{0pt}{\parskip}{0pt}
\titlespacing*{\subsection}{0pt}{\parskip}{0pt}

\usepackage{longtable,booktabs}
\usepackage{graphicx,grffile}
\usepackage{fancyhdr}

\title{Evaluating resampling methods on a real-life highly imbalanced online credit card payments dataset}

\usepackage{authblk}

\author{François de la Bourdonnaye, Fabrice Daniel}

\affil{\small Artificial Intelligence Department of Lusis, Paris, France\\http://www.lusisai.com}

\date{June 2022}
\begin{document}

\maketitle

\begin{abstract}
Various problems of any credit card fraud detection based on machine learning come from the imbalanced aspect of transaction datasets. Indeed, the number of frauds compared to the number of regular transactions is tiny and has been shown to damage learning performances, e.g., at worst, the algorithm can learn to classify all the transactions as regular. Resampling methods and cost-sensitive approaches are known to be good candidates to leverage this issue of imbalanced datasets. This paper evaluates numerous state-of-the-art resampling methods on a large real-life online credit card payments dataset. We show they are inefficient because methods are intractable or because metrics do not exhibit substantial improvements. Our work contributes to this domain in (1) that we compare many state-of-the-art resampling methods on a large-scale dataset and in (2) that we use a real-life online credit card payments dataset.
\end{abstract}

\noindent{\bf Keywords}:  resampling, gradient boosting, fraud detection 

\section{Introduction}

Credit card frauds happen daily and represent a high cost for banks and merchants, with 28.58 billion dollars of gross fraud losses in 2020 for a 2027 total amount projected at 43 billion \cite{Nilson2021}. Therefore, banks and merchants need to automatically detect a large amount of these fraudulent transactions to block them at the authorization step or raise an alert to the bank investigation team. A good hint for this is the use of supervised learning techniques, which have been known over the past years as major progresses \cite{lucas2020credit}, \cite{10.1007/978-3-319-71249-9_2}, \cite{articlepozml}. Despite these hopes, several issues related to the concept drift or imbalanced dataset are still unsolved. In this paper, we mainly focus on the problem of an imbalanced dataset. The literature addresses this issue by several methods. One well-known way consists in using cost-sensitive approaches \cite{10.5555/1642194.1642224}, \cite{cost-sensitive}. These methods consider misclassification costs in different ways, e.g., false positives can be more penalized than false negatives.
The issue of an imbalanced dataset can also be tackled by resampling the dataset such that the ratio between fraudulent and regular transactions becomes more suitable for machine learning algorithms. The two general ways of achieving it are either to withdraw some regular transactions (downsampling) \cite{1054155} or to add more fraudulent transactions (oversampling) \cite{DBLP:journals/corr/abs-1106-1813}. Note that it is also possible to combine oversampling and downsampling \cite{inproceedingsdsfsdfqqs}.

In this paper, we present various and numerous state-of-the-art resampling methods and apply them to a real-life online credit card payments dataset. The goal of these experiments is to assess whether resampling methods improve metrics or not. The contribution of our paper is related to the conducted experiments, which to our knowledge, have not been made before on a real large-scale, extremely imbalanced dataset.

We organized the remainder of the paper as follows. First, we present theoretical knowledge, including a brief presentation of gradient boosting and main resampling methods. Second, we present the experiments as well as the results. Finally, we conclude the paper.

\section{Background}
\subsection{Gradient-boosting}

\subsubsection{Main idea}
Boosting algorithms combine weak learners (generally decision trees) iteratively. Gradient boosting decision trees (GBDT) have a specific scheme where a new learner is optimized to fit the residuals of the previous learner. 

Let's describe it in mathematical terms. We use the following notations:
\begin{itemize}
    \item $F_{i} (i \in \{1,...,m\})$ is a learner
    \item $D_{\mathrm{train}} = \{(x_{1},y_{1}),...,(x_{i},y_{i}),...,(x_{n_{train}},y_{n_{train}})\} = (X,Y)$ a training dataset.
\end{itemize}

At the first iteration, $F_{1}$ is fitted as a normal decision tree, i.e., on $y$. For the following iterations $i \in \{2,...,m\}$, $F_{i}$ is fitted on $y - F_{i-1}$. Thus, the rationale is to fit estimators on the residual error of the previous estimator.

The name "gradient boosting" comes from the fact that the residual is (up to a factor) the gradient of the mean squared error between the target and the output of the learner (see Equation \ref{eqgb} and \ref{eqdgb}).

\begin{equation}
L_{\mathrm{MSE}} = (y-F_{m}(x))^{2}
 \label{eqgb}  
\end{equation}

\begin{equation}
\frac{\delta L_{\mathrm{MSE}}}{\delta F_{m}(x)} = -(y-F_{m}(x))
 \label{eqdgb}  
\end{equation}

\subsubsection{XGBoost}
XGBoost \cite{Chen_2016} is one of the most famous gradient boosting libraries. Its keys innovations are based on a novel tree learning algorithm that efficiently handles sparse data, "a theoretically justified weighted quantile sketch procedure that enables handling instance weights in approximate tree learning", the efficient use of resources for parallel computing, and "an effective cache-aware block structure or out-of-core tree learning". 

\subsubsection{LightGBM}

In conventional implementations of GBDT, the computational complexity increases with the number of features and samples. LightGBM \cite{10.5555/3294996.3295074} exhibits two techniques to overcome this difficulty: 
\begin{itemize}
    \item Gradient-based one-side sampling (GOSS): this technique under-sample data with the smallest gradients
    \item Exclusive feature bundling: this technique bundles mutually exclusive features, i.e., features that never take 0 simultaneously.
\end{itemize}
Note that we will not use GOSS for our experiments as we keep the number of features quite low, and the number of samples does not exceed ten million. 
LGBM also exhibits a built-in method to process categorical features. It sorts categories depending on the objective. Then, they use \cite{James1992} to find an optimal split according to accumulated gradients (over Hessians) in a reasonable time.

\subsubsection{CatBoost}
Catboost \cite{DBLP:journals/corr/DorogushGGKPV17}, \cite{DBLP:journals/corr/abs-1810-11363} belongs to the class of gradient boosting algorithms and is specially suitable to input spaces that contain categorical features. Two innovations are implemented:
\begin{itemize}
    \item ordered boosting
    \item algorithm that processes categorical features
\end{itemize}
The idea of ordered boosting is to compute gradients or residuals for a given sample of the training set based on a set of samples that do not include this particular training sample. Otherwise, the gradient appears to be biased. In practice, CatBoost establishes several permutations of the training set for diverse training iterations. Then, residuals are computed based on these permutations and models as well.

For processing categorical features, they take inspiration from target statistics (this converts categorical features into numerical values based on target averages). However, they proved this technique is biased and devised a solution based on ordered boosting (namely ordered TS). Categorical features can then have different values according to the training iteration.

Catboost is also innovative in a software view as it is very efficient for training and inference both in CPU and GPU.

\subsection{Resampling methods}
This section will briefly describe the main resampling methods that belong to the python package Imbalanced-learn \cite{JMLR:v18:16-365}. The user guide of the package gives more precise information about the algorithms.

\subsubsection{Down-sampling methods}

\paragraph{Random Under Sampler}
~~\\
This algorithm works as follows:
It randomly removes samples from the majority class until it reaches a hand-tuned ratio (the number of minority samples over the number of majority samples).

\paragraph{Cluster centroids}
~~\\
This algorithm uses a K-means clustering step on the majority class. It selects these K centroids to represent the majority class instead of the samples themselves.

\paragraph{NearMiss \cite{Zhang03}}
~~\\
Nearmiss presents three algorithms based on K-Nearest Neighbor algorithms:
\begin{itemize}
    \item Nearmiss-1 selects the samples from the majority class that are the closest to samples from minority classes
    \item Nearmiss-2 selects the samples from the majority class that are the farthest to samples from minority classes
    \item Nearmiss-3 is the Nearmiss-2 algorithm with a reduction step before. This step removes all the samples from the majority class that are not among the M-Nearest neighbors
\end{itemize}

\paragraph{Condensed Nearest Neighbor \cite{1054155}}
~~\\
 This algorithm works as follows:
 \begin{itemize}
     \item Gathering the minority (fraudulent transactions) samples in a set I
     \item Add a single sample from the majority class in I and the rest in a set A
     \item Scan all the elements of A and classify each sample using a 1-nearest neighbor rule
     \item If the sample is misclassified, it is added to I
     \item The scan is reiterated until no samples are added to I
    \item Then, the new reduced set to be used for classification is I
 \end{itemize}
To summarize its underlying idea, this algorithm selects samples from the majority class that are difficult to classify.

\paragraph{Tomek Links \cite{4309452}}
~~\\
First, let us define what a Tomek link is. A Tomek link exists between two samples a and b with different labels if there are mutually nearest neighbors, i.e., if the nearest neighbor of a is b and the nearest neighbor of b is a.

In practice, we withdraw either all the samples belonging to a Tomek link or just samples from the majority class belonging to a Tomek link.

\paragraph{One sided selection  \cite{1054155}}
~~\\
This algorithm derives from the Condensed Nearest Neighbor algorithm in two ways:
\begin{itemize}
    \item To avoid the presence of noisy samples, Tomek links are applied before.
    \item One iteration is applied
\end{itemize}

\paragraph{Edited Nearest Neighbor \cite{4309137}}
~~\\
This algorithm works as follows:
\begin{itemize}
    \item Scan all the samples of the majority class (normal transactions)
    \item Apply the nearest-neighbor rule for each sample
    \item Remove the sample if the selection criterion is not met.
\end{itemize}
There are two possible selection criteria: 
\begin{itemize}
    \item If the sample has all its nearest neighbors belonging to the same class, then the sample is conserved.
    \item If the sample has the majority of its nearest neighbors belonging to the same class, then the sample is conserved
\end{itemize}
This algorithm will remove majority samples close to minority samples in the feature space.

\paragraph{Repeated Edited Nearest Neighbor \cite{4309523}}
~~\\
This algorithm repeats several times the Edited Nearest Neighbor algorithm.

\paragraph{AIIKNN  \cite{4309523}}
~~\\
This algorithm is a version of Repeated Edited Nearest Neighbor that considers a higher number of nearest neighbors at each iteration.

\paragraph{Neighborhood Cleaning rule \cite{10.1007/3-540-48229-6_9}}
~~\\
This algorithm combines the Edited Nearest Neighbor algorithm (to clean the majority class) and Condensed Nearest Neighbors. 

\paragraph{Instance Hardness Threshold \cite{iht}}
~~\\
This algorithm uses a classifier (a random forest classifier by default) to fit data and remove samples from the majority classes with lower resulting probabilities.

\subsubsection{Up-sampling methods}
\paragraph{Random Over Sampler}
~~\\ 
This algorithm consists in randomly duplicating samples from the minority class.

\paragraph{SMOTE \cite{DBLP:journals/corr/abs-1106-1813}}
~~\\
This algorithm generates samples from the minority class located in a segment between a sample from the minority class and its nearest neighbors.
The algorithm steps are the following:
\begin{itemize}
    \item For each minority sample, find K nearest neighbors
    \item For each minority sample, generate new samples by interpolating the minority sample and nearest neighbors.
\end{itemize}

\paragraph{ADASYN \cite{4633969}}
~~\\
This algorithm has roughly the same idea as SMOTE, but the choice is made to sample from minority samples that have the highest number of neighbors with the majority class. The algorithm steps are the following:
\begin{itemize}
    \item Compute the total number of samples to be generated
    \item For each minority sample, find K nearest neighbors and compute a ratio between the number of nearest neighbors from the majority class and K
    \item Compute the number of samples to be generated from each minority sample (proportional to the previously computed ratio)
    \item For each minority sample, generate the required number of samples using interpolation between the sample and nearest neighbors
\end{itemize}

\paragraph{SMOTENC \cite{DBLP:journals/corr/abs-1106-1813}}
~~\\
This algorithm is a SMOTE variant that can handle categorical data without having to transform the latter. Indeed, the original SMOTE version cannot do proper interpolations on categorical data.

\paragraph{Borderline SMOTE \cite{10.1007/11538059_91}}
~~\\

This variant of SMOTE puts stress on minority samples close to the border with majority samples. This algorithm is, in practice, very close to ADASYN. The algorithm steps are the following:
\begin{itemize}
    \item Generate nearest neighbors from some minority samples
    \item For these minority samples, keep only the ones with a majority of nearest neighbors in the majority class (the samples with all its nearest neighbors from the majority class are considered noise and are also rejected).
    \item We apply a SMOTE-like step like from the conserved minority samples.
\end{itemize}

\paragraph{Kmeans SMOTE \cite{Douzas_2018}}
~~\\
This algorithm applies a KMeans clustering algorithm before applying SMOTE. After the clustering step, the minority samples from the densest clusters are used to generate new synthetic samples. The algorithm steps are the following:
\begin{itemize}
    \item Make a K-means clustering step and keep clusters with more samples from the minority class than from the majority class
 \item  For each remaining cluster, compute the sampling weight based on its
minority density
\item Oversample each filtered cluster using SMOTE and sampling weight to compute the number of samples to be generated 
\end{itemize}

\paragraph{SVM SMOTE \cite{10.1504/IJKESDP.2011.039875}}
~~\\

This algorithm generates samples from support vectors of an SVM classifier. The algorithm steps are the following:
\begin{itemize}
    \item Train an SVM classifier on training data and generate support vectors of the minority class
    \item Generate nearest neighbors of minority support vectors
    \item Create new minority samples by using interpolation between minority support vectors and nearest neighbors (note that the algorithm handles the case in which the majority of nearest neighbors come from majority samples and act accordingly)
\end{itemize}

\subsubsection{Combined methods}
We also test combined methods such as SMOTENN and SMOTETOMEK. They are nothing but the application of SMOTE followed by the application of ENN (or TOMEK).

\section{Experiments}

\subsection{Settings}

\subsubsection{Datasets}

We use a subset of 10 million transactions from real-life data of a major French bank using raw and derivative features. The raw features consist of 8 categorical features and 2 numerical features. The derivative features are 20 numerical features computed using count, mean, and diff aggregators.

\subsubsection{Methodology}
\paragraph{Tractability study}
~~\\
We use subsets of the mentioned above dataset of different sizes. The experiments happen as follows:
\begin{itemize}
    \item We train each resampling method on these subsets
    \item We apply linear, logarithmic, polynomial, and exponential regressions
    \item We select the regression method that gives the best r2 score (coefficient of determination)
    \item With the fitted regressors, we predict resampling time for the training set (6 700 000 transactions)
    \item We select the resampling method that presents reasonable predicted resampling times and rejects the other ones
\end{itemize}

\paragraph{Comparison of resampling methods}
~~\\
All this paragraph applies to methods that presented reasonable predicted resampling times.

We split the datasets into training and validation sets according to a $\frac{2}{3}/ \frac{1}{3}$ policy.
Then, for each gradient boosting method, we apply the CatBoost encoder \cite{DBLP:journals/corr/abs-1810-11363} we have found to be the best tradeoff for these models in a previous study \cite{DBLP:journals/corr/abs-2112-12024}.

After that, we optimize the hyperparameters of these gradient boosting models using the following approach: 

The optimization happened using 3-fold cross-validation on the training set (the optimization criterion was the PR AUC) using the Optuna package \cite{optuna_2019}, and the Tree-Parzen-Estimator as the sampling strategy \cite{NIPS2011_86e8f7ab}. 

We resample the training set using selected resampling methods.
For each resampled dataset, we apply each gradient boosting model (LGBM, XGBoost, CatBoost). Then we evaluate the methods using the following metrics:
\begin{itemize}
    \item Area under Precision-recall curve (PR AUC )
    \item Precision
    \item Recall
    \item F1
\end{itemize}
As ranks between methods are not stable over the seeds given as input to the boosting models, we average each setting over ten seeds. We do not use the accuracy metric since it does not make much sense with a highly imbalanced dataset.

\subsubsection{Hyperparameters for resampling methods}
We use the default values for all the methods except for the "sampling$\_$strategy" parameter we set to 0.1 when it exists. The methods are the following:
\begin{itemize}
    \item Cluster Centroids
    \item SVM SMOTE
    \item SMOTE ENN
    \item SMOTE TOMEK
    \item Instance Near Miss
    \item Instance Hardness Threshold
    \item SMOTE
    \item SMOTE-NC
    \item Random over sampler
    \item ADASYN
    \item BORDERLINE SMOTE
    \item Random under sampler
\end{itemize}
\subsection{Results}

\subsubsection{Tractability study}

We rejected 11 methods over 19 because of too high predicted resampling time. Table \ref{table_resampling_time} lists the resampling methods, predicted computation time, and the decision to reject it or not.

\begin{table}[h!]
    \centering
    
    \begin{tabular}{|p{4cm}|p{1.5cm}|p{1.3cm}|}
    \hline
         Resampling method & Predicted computation time & Decision \\
         \hline
         Cluster Centroids &$>$11 weeks &Rejected \\
         \hline
         Condensed Nearest Neighbour & $>$489 weeks &Rejected \\
         \hline
          Edited Nearest Neighbour &$>$2 weeks &Rejected \\
         \hline
          Repeated Edited Nearest Neighbour & $>$2 weeks &Rejected \\
         \hline
          ALL KNN & $>$5 weeks & Rejected\\
         \hline
          Tomek link & $>$ 1 week & Rejected\\
         \hline
        One Sided Selection & $>$ 1 week & Rejected\\
         \hline
        SVM SMOTE &1 day, 2 hours, 35 minutes, 46 seconds & Rejected\\
         \hline
        SMOTE ENN &$>$4 weeks & Rejected\\
         \hline
        SMOTE Tomek &$>$2 weeks & Rejected\\
         \hline
        Instance neighborhood cleaning rule & $>$2 weeks &Rejected\\
         \hline
        Instance near miss & 14 minutes, 14 seconds &Accepted\\
         \hline
        Instance hardness threshold & 2 minutes, 1 second &Accepted\\
         \hline
        SMOTE & 21 seconds & Accepted\\
         \hline
        SMOTE-NC & 9 minutes, 13 seconds& Accepted\\
         \hline
        Random over sampler & 7 seconds & Accepted\\
         \hline
        Adasyn & 58 minutes, 55 seconds& Accepted\\
         \hline
        Borderline SMOTE & 1h, 18 minutes, 58 seconds & Accepted\\
         \hline
        Random under sampler & 3 seconds & Accepted\\
         \hline
    \end{tabular}
    \caption{Results of tractability study}
    \label{table_resampling_time}
\end{table}

\newpage

We have rejected methods that have a resampling time exceeding one day.
In practice, we found that predicted times were roughly twice lower for tractable resampling methods than actual computation times.

We performed these predictions and all the computations in this paper on the Mésocentre of CentraleSupelec managed by IDRIS (CNRS), used in the context of Lusis/CentraleSupelec AI research chair. Our computing allocates 80 hyperthreaded cores Xeon Gold 6178 at 2.4GHz with 1TB memory.

\subsubsection{Comparison Results of tractable resampling methods}
For each resampling/gradient boosting model pair, we display the metrics by using a table gathering the average results of each method. More precisely, we only display the percentage of variations compared to the absence of resampling. We willingly omit raw values for confidentiality purposes.

\textbf{LGBM}

\begin{table}[h!]
    \centering
    \begin{tabular}{|p{1.4cm}|c|c|c|c|}
    \hline
         &  PR AUC & Precision & Recall & F1 \\
    \hline
    Instance Hardness Threshold    & -25\% & -77\% & +525\% & +64\% \\
     \hline
    Random under sampler   & \textbf{-7\%} & -76\% & +531\%& +73\% \\
     \hline
    Random over sampler& \textbf{-7\%} & \textbf{-71\%} & +462\% & \textbf{+90\%} \\
     \hline
    Borderline SMOTE    & -38\% & -77\% & +356\% & +48\% \\
     \hline
    SMOTE    & -59\% & -91\% & +527\% & -26\% \\
     \hline
    ADASYN    & -68\% & -93\% & +606\% & -40\% \\
     \hline
    NEAR MISS    & -97\% & -99\% & \textbf{+1293\%} & -93\% \\
     \hline
    \end{tabular}
    \caption{Results for LGBM}
    \label{lgbm_res}
\end{table}

We establish several observations:
\begin{itemize}
    \item We observe that random undersampling, oversampling, instance hardness threshold and borderline SMOTE method improve recall and F1 while deteriorating Precision Recall AUC (PR AUC) and Precision.
    \item We observe that other methods only improve the recall metric and significantly damage the other ones.
    \item Then, overall, we observe that the metrics are not substantially improved.
\end{itemize}
    
\textbf{CatBoost}

\begin{table}[h!]
    \centering
    \begin{tabular}{|p{1.4cm}|c|c|c|c|}
    \hline
         &  PR AUC & Precision & Recall & F1 \\
    \hline
    Instance Hardness Threshold    & -32\% & -85\% & +782\% & +45\% \\
     \hline
    Random under sampler   & \textbf{-18\%} & -84\% & +757\% & +57\% \\
     \hline
    Random over sampler& -27\% & \textbf{-75\%} & +481\% & \textbf{+96\%} \\
     \hline
    Borderline SMOTE    & -48\% & -81\% & +406\% & +51\% \\
     \hline
    SMOTE    & -64\% & -92\% & +623\% & -16\% \\
     \hline
    ADASYN    & -65\% & -93\% & +769\% & -24\% \\
     \hline
    NEAR MISS    & -97\% & -99\% & \textbf{+1602\%} & -91\% \\
     \hline
    \end{tabular}
    \caption{Results for CatBoost}
    \label{lgbm_CatBoost}
\end{table}
We establish same observations as before:
\begin{itemize}
    \item We observe that random undersampling, oversampling, instance hardness threshold and borderline SMOTE method improve recall and F1 while deteriorating PR AUC and Precision.
    \item We observe that other methods only improve the recall metric and significantly damage the other ones.
    \item Then, overall, we observe that the metrics are not substantially improved.
\end{itemize}

\textbf{XGBoost}

\begin{table}[h!]
    \centering
    \begin{tabular}{|p{1.4cm}|c|c|c|c|}
    \hline
         &  PR AUC & Precision & Recall & F1 \\
    \hline
    Instance Hardness Threshold    & -31\% & -85\% & +841\% & +46\% \\
     \hline
    Random under sampler   & \textbf{-24\%} & -84\% & +801\% & +55\% \\
     \hline
    Random over sampler& -25\% & \textbf{-78\%} & +606\% & \textbf{+95\%} \\
     \hline
    Borderline SMOTE    & -45\% & -80\% & +448\% & +70\% \\
     \hline
    SMOTE    & -67\% & -92\% & +707\% & -18\% \\
     \hline
    ADASYN    & -72\% & -93\% & +836\% & -28\% \\
     \hline
    NEAR MISS    & -97\% & -99\% & \textbf{+1657\%} & -91\% \\
     \hline
    \end{tabular}
    \caption{Results for XGBoost}
    \label{lgbm_XGBoost}
\end{table}
We establish roughly the same observations as before:
\begin{itemize}
    \item We observe that random undersampling, oversampling, instance hardness threshold and borderline smote methods improve recall and F1 while deteriorating PR AUC and Precision.
    \item We observe that other methods only improve the recall metric and significantly damage the other ones.
    \item Then, overall, we observe that the metrics are not substantially improved.
\end{itemize}
\section{Conclusion}

We first conducted a computation time study in which the goal was to assess which method was tractable for a large-scale dataset. It has allowed us to remove resampling methods we cannot compute in a reasonable time, e.g., among the downsampling methods, only Instance Hardness Threshold and Instance Near miss were considered tractable.
After that, for any selected method, recall remarkably increased, but precision decreased significantly. We can explain the recall increase by resampling itself. However, the significant precision decrease may be due to a loss of helpful information for undersampling methods. It can also be due to the generation of useless information for oversampling methods. Furthermore, these observations seem in agreement with the literature.

Consequently, resampling techniques are unsuitable for overcoming imbalanced aspects of our large-scale and real-life online credit card payments dataset.

Gradient Boosting alone is giving better results. As shown in \cite{freryEnsembleLearningExtremely2019} Gradient Boosting focus on hard examples. The algorithm is driven by the minority class when the class distribution is highly imbalanced.

Thus, as hints for future work, we wish to focus on Gradient Boosting, cost-sensitive learning \cite{10.5555/1642194.1642224} or machine learning methods that naturally integrate techniques to deal with imbalanced aspects \cite{inproceedingsefsdfdsf}.

\newpage

\printbibliography

@article{Chen_2016,
   title={XGBoost},
   ISBN={9781450342322},
   url={http://dx.doi.org/10.1145/2939672.2939785},
   DOI={10.1145/2939672.2939785},
   journal={Proceedings of the 22nd ACM SIGKDD International Conference on Knowledge Discovery and Data Mining},
   publisher={ACM},
   author={Chen, Tianqi and Guestrin, Carlos},
   year={2016},
   month={Aug}
}

@inproceedings{10.5555/3294996.3295074,
author = {Ke, Guolin and Meng, Qi and Finley, Thomas and Wang, Taifeng and Chen, Wei and Ma, Weidong and Ye, Qiwei and Liu, Tie-Yan},
title = {LightGBM: A Highly Efficient Gradient Boosting Decision Tree},
year = {2017},
isbn = {9781510860964},
publisher = {Curran Associates Inc.},
address = {Red Hook, NY, USA},
booktitle = {Proceedings of the 31st International Conference on Neural Information Processing Systems},
pages = {3149–3157},
numpages = {9},
location = {Long Beach, California, USA},
series = {NIPS'17}
}

@article{DBLP:journals/corr/DorogushGGKPV17,
  author    = {Anna Veronika Dorogush and
               Andrey Gulin and
               Gleb Gusev and
               Nikita Kazeev and
               Liudmila Ostroumova Prokhorenkova and
               Aleksandr Vorobev},
  title     = {CatBoost: unbiased boosting with categorical features},
  journal   = {CoRR},
  volume    = {abs/1706.09516},
  year      = {2017},
  url       = {http://arxiv.org/abs/1706.09516},
  archivePrefix = {arXiv},
  eprint    = {1706.09516},
  bibsource = {dblp computer science bibliography, https://dblp.org}
}

@article{DBLP:journals/corr/abs-1810-11363,
  author    = {Anna Veronika Dorogush and
               Vasily Ershov and
               Andrey Gulin},
  title     = {CatBoost: gradient boosting with categorical features support},
  journal   = {CoRR},
  volume    = {abs/1810.11363},
  year      = {2018},
  url       = {http://arxiv.org/abs/1810.11363},
  archivePrefix = {arXiv},
  eprint    = {1810.11363},
  bibsource = {dblp computer science bibliography, https://dblp.org}
}

@Inbook{James1992,
author="James, W.
and Stein, Charles",
editor="Kotz, Samuel
and Johnson, Norman L.",
title="Estimation with Quadratic Loss",
bookTitle="Breakthroughs in Statistics: Foundations and Basic Theory",
year="1992",
publisher="Springer New York",
address="New York, NY",
pages="443--460",
isbn="978-1-4612-0919-5",
doi="10.1007/978-1-4612-0919-5_30",
url="https://doi.org/10.1007/978-1-4612-0919-5_30"
}

@article{articlepozml,
author = {Dal Pozzolo, Andrea and Boracchi, Giacomo and Caelen, Olivier and Alippi, Cesare and Bontempi, Gianluca},
year = {2017},
month = {09},
pages = {1-14},
title = {Credit Card Fraud Detection: A Realistic Modeling and a Novel Learning Strategy},
volume = {PP},
journal = {IEEE Transactions on Neural Networks and Learning Systems},
doi = {10.1109/TNNLS.2017.2736643}
}

@InProceedings{10.1007/978-3-319-71249-9_2,
author="Frery, Jordan
and Habrard, Amaury
and Sebban, Marc
and Caelen, Olivier
and He-Guelton, Liyun",
editor="Ceci, Michelangelo
and Hollm{\'e}n, Jaakko
and Todorovski, Ljup{\v{c}}o
and Vens, Celine
and D{\v{z}}eroski, Sa{\v{s}}o",
title="Efficient Top Rank Optimization with Gradient Boosting for Supervised Anomaly Detection",
booktitle="Machine Learning and Knowledge Discovery in Databases",
year="2017",
publisher="Springer International Publishing",
address="Cham",
pages="20--35",
isbn="978-3-319-71249-9"
}

@misc{lucas2020credit,
      title={Credit card fraud detection using machine learning: A survey}, 
      author={Yvan Lucas and Johannes Jurgovsky},
      year={2020},
      eprint={2010.06479},
      archivePrefix={arXiv},
      primaryClass={cs.LG}
}

@article{cost-sensitive,
author = {Ling, Charles and Sheng, Victor},
year = {2010},
month = {01},
pages = {},
title = {Cost-Sensitive Learning and the Class Imbalance Problem},
journal = {Encyclopedia of Machine Learning}
}

@inproceedings{10.5555/1642194.1642224,
author = {Elkan, Charles},
title = {The Foundations of Cost-Sensitive Learning},
year = {2001},
isbn = {1558608125},
publisher = {Morgan Kaufmann Publishers Inc.},
address = {San Francisco, CA, USA},
booktitle = {Proceedings of the 17th International Joint Conference on Artificial Intelligence - Volume 2},
pages = {973–978},
numpages = {6},
location = {Seattle, WA, USA},
series = {IJCAI'01}
}

@article{DBLP:journals/corr/abs-1106-1813,
  author    = {Kevin W. Bowyer and
               Nitesh V. Chawla and
               Lawrence O. Hall and
               W. Philip Kegelmeyer},
  title     = {{SMOTE:} Synthetic Minority Over-sampling Technique},
  journal   = {CoRR},
  volume    = {abs/1106.1813},
  year      = {2011},
  url       = {http://arxiv.org/abs/1106.1813},
  eprinttype = {arXiv},
  eprint    = {1106.1813},
  bibsource = {dblp computer science bibliography, https://dblp.org}
}

@ARTICLE{1054155,

  author={Hart, P.},

  journal={IEEE Transactions on Information Theory}, 

  title={The condensed nearest neighbor rule (Corresp.)}, 

  year={1968},

  volume={14},

  number={3},

  pages={515-516},

  doi={10.1109/TIT.1968.1054155}}

@article{JMLR:v18:16-365,
author  = {Guillaume  Lema{{\^i}}tre and Fernando Nogueira and Christos K. Aridas},
title   = {Imbalanced-learn: A Python Toolbox to Tackle the Curse of Imbalanced Datasets in Machine Learning},
journal = {Journal of Machine Learning Research},
year    = {2017},
volume  = {18},
number  = {17},
pages   = {1-5},
url     = {http://jmlr.org/papers/v18/16-365.html}
}

@ARTICLE{4309137,

  author={Wilson, Dennis L.},

  journal={IEEE Transactions on Systems, Man, and Cybernetics}, 

  title={Asymptotic Properties of Nearest Neighbor Rules Using Edited Data}, 

  year={1972},

  volume={SMC-2},

  number={3},

  pages={408-421},

  doi={10.1109/TSMC.1972.4309137}}

@ARTICLE{4309523,  author={},  journal={IEEE Transactions on Systems, Man, and Cybernetics},   title={An Experiment with the Edited Nearest-Neighbor Rule},   year={1976},  volume={SMC-6},  number={6},  pages={448-452},  doi={10.1109/TSMC.1976.4309523}}

@ARTICLE{4309452,  author={},  journal={IEEE Transactions on Systems, Man, and Cybernetics},   title={Two Modifications of CNN},   year={1976},  volume={SMC-6},  number={11},  pages={769-772},  doi={10.1109/TSMC.1976.4309452}}

@InProceedings{10.1007/3-540-48229-6_9,
author="Laurikkala, Jorma",
editor="Quaglini, Silvana
and Barahona, Pedro
and Andreassen, Steen",
title="Improving Identification of Difficult Small Classes by Balancing Class Distribution",
booktitle="Artificial Intelligence in Medicine",
year="2001",
publisher="Springer Berlin Heidelberg",
address="Berlin, Heidelberg",
pages="63--66",
isbn="978-3-540-48229-1"
}

@article{iht,
author = {Smith, Michael and Martinez, Tony and Giraud-Carrier, Christophe},
year = {2013},
month = {11},
pages = {},
title = {An Instance Level Analysis of Data Complexity},
journal = {Machine Learning},
doi = {10.1007/s10994-013-5422-z}
}

@inproceedings{Zhang03,
  author = {Zhang, J. and Mani, I.},
  booktitle = {{Proceedings of the ICML'2003 Workshop on Learning from Imbalanced Datasets}},
  description = {The big one},
  title = {{KNN Approach to Unbalanced Data Distributions: A Case Study Involving Information Extraction}},
  year = 2003
}

@InProceedings{10.1007/11538059_91,
author="Han, Hui
and Wang, Wen-Yuan
and Mao, Bing-Huan",
editor="Huang, De-Shuang
and Zhang, Xiao-Ping
and Huang, Guang-Bin",
title="Borderline-SMOTE: A New Over-Sampling Method in Imbalanced Data Sets Learning",
booktitle="Advances in Intelligent Computing",
year="2005",
publisher="Springer Berlin Heidelberg",
address="Berlin, Heidelberg",
pages="878--887"
}

@INPROCEEDINGS{4633969,  author={Haibo He and Yang Bai and Garcia, Edwardo A. and Shutao Li},  booktitle={2008 IEEE International Joint Conference on Neural Networks (IEEE World Congress on Computational Intelligence)},   title={ADASYN: Adaptive synthetic sampling approach for imbalanced learning},   year={2008},  volume={},  number={},  pages={1322-1328},  doi={10.1109/IJCNN.2008.4633969}}

@article{10.1504/IJKESDP.2011.039875,
author = {Nguyen, Hien M. and Cooper, Eric W. and Kamei, Katsuari},
title = {Borderline Over-Sampling for Imbalanced Data Classification},
year = {2011},
issue_date = {April 2011},
publisher = {Inderscience Publishers},
address = {Geneva 15, CHE},
volume = {3},
number = {1},
issn = {1755-3210},
url = {https://doi.org/10.1504/IJKESDP.2011.039875},
doi = {10.1504/IJKESDP.2011.039875},
journal = {Int. J. Knowl. Eng. Soft Data Paradigm.},
month = apr,
pages = {4–21},
numpages = {18}
}

@article{Douzas_2018,
   title={Improving imbalanced learning through a heuristic oversampling method based on k-means and SMOTE},
   volume={465},
   ISSN={0020-0255},
   url={http://dx.doi.org/10.1016/j.ins.2018.06.056},
   DOI={10.1016/j.ins.2018.06.056},
   journal={Information Sciences},
   publisher={Elsevier BV},
   author={Douzas, Georgios and Bacao, Fernando and Last, Felix},
   year={2018},
   month={Oct},
   pages={1–20}
}

@inproceedings{inproceedingsefsdfdsf,
author = {Hido, Shohei and Kashima, Hisashi},
year = {2008},
month = {04},
pages = {143-152},
title = {Roughly Balanced Bagging for Imbalanced Data},
doi = {10.1137/1.9781611972788.13}
}

@inproceedings{inproceedingsdsfsdfqqs,
author = {Batista, Gustavo and Bazzan, Ana and Monard, Maria-Carolina},
year = {2003},
month = {01},
pages = {10-18},
title = {Balancing Training Data for Automated Annotation of Keywords: a Case Study.},
journal = {the Proc. Of Workshop on Bioinformatics}
}

@inproceedings{optuna_2019,
    title={Optuna: A Next-generation Hyperparameter Optimization Framework},
    author={Akiba, Takuya and Sano, Shotaro and Yanase, Toshihiko and Ohta, Takeru and Koyama, Masanori},
    booktitle={Proceedings of the 25rd {ACM} {SIGKDD} International Conference on Knowledge Discovery and Data Mining},
    year={2019}
}

@inproceedings{NIPS2011_86e8f7ab,
 author = {Bergstra, James and Bardenet, R\'{e}mi and Bengio, Yoshua and K\'{e}gl, Bal\'{a}zs},
 booktitle = {Advances in Neural Information Processing Systems},
 editor = {J. Shawe-Taylor and R. Zemel and P. Bartlett and F. Pereira and K. Q. Weinberger},
 pages = {},
 publisher = {Curran Associates, Inc.},
 title = {Algorithms for Hyper-Parameter Optimization},
 url = {https://proceedings.neurips.cc/paper/2011/file/86e8f7ab32cfd12577bc2619bc635690-Paper.pdf},
 volume = {24},
 year = {2011}
}

@article{freryEnsembleLearningExtremely2019,
  title = {Ensemble {{Learning}} for {{Extremely Imbalanced Data Flows}}},
  author = {Frery, Jordan},
  year = {2019},
  pages = {150},
  language = {en}
}

@techreport{Nilson2021,
  title       = {Nilson Report, issue 1209},
  institution = {Nilson Report},
  year        = {2021},
  month       = {dec}
}

@article{DBLP:journals/corr/abs-2112-12024,
  author    = {Fran{\c{c}}ois De La Bourdonnaye and
               Fabrice Daniel},
  title     = {Evaluating categorical encoding methods on a real credit card fraud
               detection database},
  journal   = {CoRR},
  volume    = {abs/2112.12024},
  year      = {2021},
  url       = {https://arxiv.org/abs/2112.12024},
  eprinttype = {arXiv},
  eprint    = {2112.12024},
  bibsource = {dblp computer science bibliography, https://dblp.org}
}

\end{document}